\documentclass{article}
\usepackage{ijcai20}

\usepackage{times}

\usepackage{soul}
\usepackage{url}
\usepackage[hidelinks]{hyperref}
\usepackage[utf8]{inputenc}
\usepackage[small]{caption}
\usepackage{graphicx}
\usepackage{amsmath}
\usepackage{booktabs}
\usepackage{natbib}
\usepackage{amssymb}
\usepackage{mathtools}

\DeclareMathOperator*{\argmax}{arg\,max}

\title{Think Too Fast Nor Too Slow: \\ The Computational Trade-off Between Planning And Reinforcement Learning}

\author{
Thomas M. Moerland$^{1,2}$\footnote{Authors contributed equally.} \and 
Anna Deichler$^{1,4*}$ \and  
Simone Baldi$^{3,4}$ \and 
Joost Broekens$^{2}$ \And \\
Catholijn M. Jonker$^{1,2}$ 
\affiliations
$^1$Interactive Intelligence, TU Delft, The Netherlands\\
$^2$Leiden Institute of Advanced Computer Science, Leiden University, The Netherlands\\
$^3$School of Cyberscience and Engineering, Southeast University, China\\
$^4$Delft Center for Systems and Control, TU Delft, The Netherlands\\
\emails
T.M.Moerland@tudelft.nl
}

\begin{document}

\maketitle

\begin{abstract}
Planning and reinforcement learning are two key approaches to sequential decision making. Multi-step approximate real-time dynamic programming, a recently successful algorithm class of which AlphaZero \citep{silver2018general} is an example, combines both by nesting planning within a learning loop. However, the combination of planning and learning introduces a new question: how should we balance time spend on planning, learning and acting? The importance of this trade-off has not been explicitly studied before. We show that it is actually of key importance, with computational results indicating that we should neither plan too long nor too short. Conceptually, we identify a new spectrum of planning-learning algorithms which ranges from exhaustive search (long planning) to model-free RL (no planning), with optimal performance achieved midway.  
\end{abstract}

\section{Introduction}
Sequential decision-making, commonly formalized as Markov Decision Process (MDP) optimization, is a key challenge in artificial intelligence (AI) and machine learning research. Important solution approaches include planning (or search) \citep{russell2016artificial} and reinforcement learning \citep{sutton2018reinforcement}. Recently, a class of algorithms, known as multi-step approximate real-time dynamic programming (MSA-RTDP), combines both fields. MSA-RTDP iterates planning, which uses a learned value/policy function, and learning, which uses output from the planning procedure. A successful example in this class is the AlphaZero algorithm, which achieved super-human performance in the game of Go, Chess, and Shogi \citep{silver2017mastering,silver2018general}. 

This iterated planning and learning procedure introduces a crucial new question: how long should we plan at a given state? We hypothesize that this is a crucial trade-off for planning-learning integrations: when we plan too extensively, we make too little progress in the domain and have less training targets for learning, while when we plan too briefly, our local decisions and training targets are likely to be less optimal. This trade-off was never present in online planning, where the budget per real step is typically as high as the application permits (in the order of milliseconds for a video game, or in the order of seconds to minutes for a game of Chess \citep{campbell2002deep}). It was neither present in model-free reinforcement learning (RL), since those approaches do not have access to a dynamics model and can therefore not plan. Model-based RL, where we use observed data to approximate the dynamics model, has mostly focused on dealing with enhancing data efficiency and dealing with uncertainty in the learned models \citep{sutton1991dyna,chua2018deep}. Instead, we focus on the situation with a known, perfect model without uncertainty, to fully investigate the trade-off between planning and learning once a good model is available.

We therefore study the AlphaZero algorithm on several known tasks, where we fix the overall computational budget, but vary the planning budget per real step and associated training iteration. Our results show that, for a fixed overall time budget, approaches with an intermediate planning budget per time-step achieve the highest final performance. First, this is an important empirical insight for model-based reinforcement learning and MSA-RTDP algorithms. Moreover, the fundamental mutual benefit of planning and learning, which outperforms their isolated application, may also provide an argument for the existence of fast prediction (System 1) and explicit planning (System 2) in human decision making. This theory, better known as dual process theory \citep{evans1984heuristic}, was more recently popularized as `thinking fast and slow' \citep{kahneman2011thinking}. A short summary of our results could be: `think too fast nor too slow'.

The remainder of this paper is organized as follows. Section \ref{sec_preliminaries} provides essential background on Markov Decision Process optimization, while Section \ref{sec_rtdp} introduces the algorithm class of interest, multi-step approximate real-time dynamic programming. Section \ref{sec_methods} and \ref{sec_results} detail methodology and results, respectively. The final sections cover Related work (Sec. \ref{sec_related_work}), Discussion (Sec. \ref{sec_discussion}) and Conclusion (Sec. \ref{sec_conclusion}). Code to replicate experiments is available from \emph{\url{https://github.com/ratponto/tree-rl-adaptive}}.

\section{Preliminaries} \label{sec_preliminaries}
We study the {\it Markov Decision Process} (MDP) \citep{puterman2014markov} optimization problem. An MDP is defined by a state space $\mathcal{S}$, an action space $\mathcal{A}$, a transition function $T: \mathcal{S} \times \mathcal{A} \to p(\mathcal{S})$, a reward function $\mathcal{R}: \mathcal{S} \times \mathcal{A} \times \mathcal{S} \to \mathbb{R}$, an initial state distribution $p(s_0)$ and a discount parameter $\gamma \in [0,1]$. 

We can interact with the environment through a policy $\pi: \mathcal{S} \to p(\mathcal{A})$. After specifying an action $a_t$ in state $s_t$, the environment returns a next state $s_{t+1} \sim \mathcal{T}(\cdot|s_t,a_t)$ and associated reward $r_t = \mathcal{R}(s_t,a_t,s_{t+1})$. We are interested in finding the policy that gives the highest cumulative pay-off. Define the state-action value as:

\begin{equation} 
Q(s,a) \dot{=} \mathbb{E}_{\pi,\mathcal{T}} \Bigg[ \sum_{k=0}^K \gamma^k r_{t+k}  \Big| s_t=s, a_t=a \Bigg] \label{eq_Q}
\end{equation}

and $V(s) = \mathbb{E}_{a\sim\pi(\cdot|s)} [Q(s,a)]$. There is only one optimal value function $Q^\star(s,a)$ \citep{sutton2018reinforcement}, and our goal is to find an optimal policy $\pi^\star$ that achieves the optimal value:

\begin{equation}
\pi^\star = \argmax_\pi Q(s,a). \label{eq_pi_star}
\end{equation} 

The possible approaches to this problem crucially rely on our type of access to the environment dynamics $\mathcal{T}$ and reward function $\mathcal{R}$. In model-free reinforcement learning, the environment cannot be reverted, and we therefore have to sample forward from the state that we reach. This property, also referred to as an `unknown model', is also part of the real world. In contrast, in planning and model-based RL, we are either given or have learned a reversible model, better known as a `known model', which we can query for a next state and reward for any state-action pair that we impute.

A classic approach in the latter case (known model) is Dynamic Programming (DP) \citep{bellman1966dynamic}. For example, in Q-value iteration we sweep through a state-action value table, where at each location we update $Q(s,a)$ according to: 

\begin{equation}
Q(s,a) \gets  \mathbb{E}_{s' \sim \mathcal{T}(\cdot|s,a)} \Big[ \mathcal{R}(s,a,s') + \gamma \max_{a \in \mathcal{A}} Q(s,a) \Big]
\end{equation}

Dynamic programming is guaranteed to converge to the optimal policy. However, due to the curse of dimensionality, it can not be applied in high-dimensional problems. In the next section we introduce a recently popularized extension of DP. 

\section{Multi-step Approximate Real-Time Dynamic Programming} \label{sec_rtdp}
Multi-step approximate real-time dynamic programming \citep{efroni2019multi} has recently shown impressive empirical results, for example beating humans and achieving state-of-the-art performance in the game of Go \citep{silver2017mastering}, Chess and Shogi \citep{silver2018general}. MSA-RTDP is based on Dynamic Programming concepts, but adds three additional concepts:

\begin{itemize}
\item `Real time' \citep{barto1995learning} implies that we act on traces through the environment that start from some initial state $s_0 \sim p(s_0)$. This property is assumed by most RL and planning algorithms. Compared to the DP sweeps, it avoids work on states that we will never reach. 
\item `Approximate' implies that we will use function approximation to store a global parametrized solution, in the form of a value $V_\theta(s)$/$Q_\theta(s,a)$ and/or policy function $\pi_\theta(a|s)$, where $\theta \in \Theta$ denote the parameters of the approximation. Compared to a tabular representation, approximate representations can deal with high-dimensional state spaces and benefit from generalization between similar states, although they do make approximation errors. Approximate solutions are especially popular in RL literature.  
\item `Multi step' implies that for every Dynamic Programming back-up, we are allowed to make a multi-step lookahead, i.e., we can {\it plan}.  
\end{itemize}

\begin{figure}[!t]
  \centering
      \includegraphics[width = 0.49\textwidth]{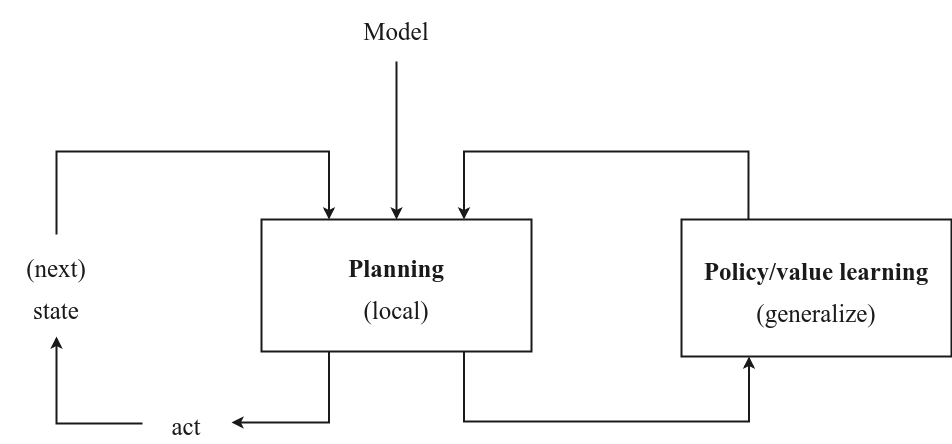}
  \caption{Multi-step Real-time Dynamic Programming. The three key procedures are 1) Planning, 2) Learning, and 3) Real steps (acting).}
    \label{fig_rtdp}
\end{figure} 

The resulting multi-step approximate RTDP algorithm class has three key components, which are visualized in Figure \ref{fig_rtdp}:
\begin{enumerate}
\item {\bf Plan}: At every state $s_t$ in the trace, we get to expand some computational budget $\mathtt{B}$ of forward planning, which could for example be a depth-$d$ full-breadth search \citep{russell2016artificial}, or a more complicated planning procedure like Monte Carlo Tree Search \citep{browne2012survey}. The planning procedure can use learned value/policy functions to aid planning, for example through {\it bootstrapping} \citep{sutton2018reinforcement}.     
\item {\bf Learn}: After planning, we use the output of planning (our improved knowledge about the optimal value and policy at $s_t$) to train our global value/policy approximation. 
\item {\bf Real step}: We finally use the planning output to decide which action $a_t$ we will commit to, and make a `real step', transitioning to a sampled next state $s_{t+1} \sim \mathcal{T}(\cdot|s_t,a_t)$. The next iteration of planning continues from $s_{t+1}$.
\end{enumerate}

\begin{figure*}[h]
  \centering
      \includegraphics[width = 0.90\textwidth]{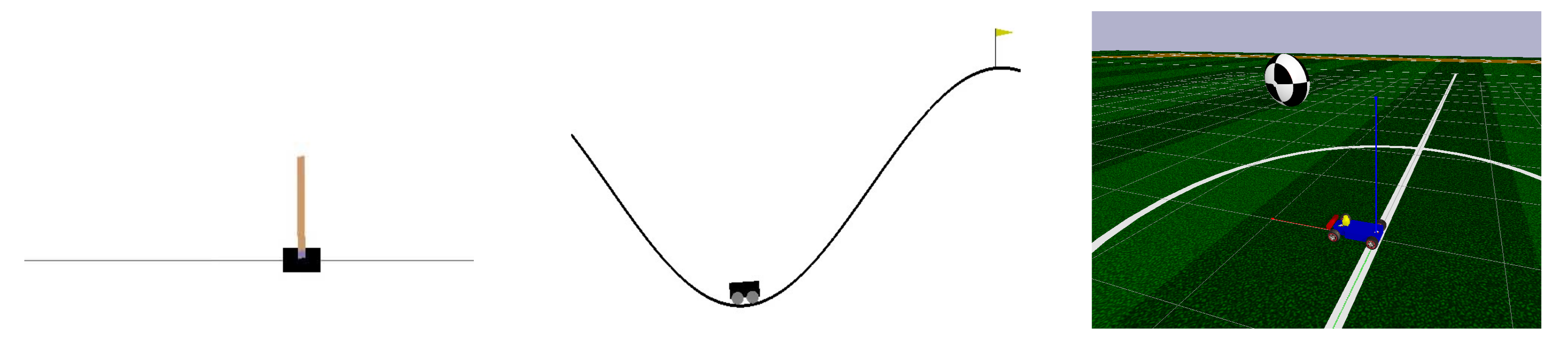}
  \caption{Image stills from the studied tasks. Left: CartPole, where we attempt to balance the pole. Middle: MountainCar, where we attempt to reach the top-left flag by swinging back and forth. Right: RaceCar, where we need to control a car to reach a goal, indicated by a ball.}
    \label{fig_tasks}
\end{figure*} 

MSA-RTDP has two special cases that depend on the computation planning budget $\mathtt{B}$ per real step. One the one extreme, $\mathtt{B} \to \infty$, we completely enumerate all possible future traces, better known as {\it exhaustive search} \citep{russell2016artificial}. On the other extreme, $\mathtt{B} = 0$, we do not plan at all, but directly make a real step based on the global approximations, better known as {\it model-free reinforcement learning} \citep{sutton2018reinforcement}.

\citet{anthony2017thinking} already related this approach to cognitive psychology research, in particular dual process theory \citep{evans1984heuristic,kahneman2011thinking}. The global value/policy approximation, which makes fast predictions about the value of actions, can be considered a System 1 (`Thinking Fast'), while explicit forward planning to improve over these fast approximations seems related to System 2 (`Thinking slow').

\section{Methods} \label{sec_methods}
For this paper, we will follow the AlphaGo Zero \citep{silver2017mastering} variant of MSA-RTDP. AlphaGo Zero uses a variant of MCTS \citep{browne2012survey} for planning, and deep neural networks for leaning of a policy $\pi_\theta(a|s)$ and value $V_\theta(s)$ approximation. A key aspect of iterated planning-learning is their mutual influence, where planning improves the learned function, and the learned function directs new planning iterations. We will detail both these integrations, starting with training target construction based on planning output. 

To train the policy network, we normalize the action visitation counts $n(s,a)$ at the tree root state $s$ to a probability distribution, and train on a cross-entropy loss: 

\begin{equation}
\text{L}_\pi(\theta) = \sum_a \frac{n(s,a)}{n(s)} \log \pi_\theta(a|s). 
\end{equation}

For value network training, we use a target based on the reweighted value estimates at the root of the MCTS,

\begin{equation} 
\hat{V}(s) = \sum_a \frac{n(s,a)}{n(s)} \bar{Q}(s,a),
\end{equation}

where $\bar{Q}(s,a)$ denotes the mean pay-off of all traces through $(s,a)$, and train on a squared error loss,

\begin{equation}
\text{L}_V(\theta) = \big( V_\theta(s) - \hat{V}(s) \big)^2. 
\end{equation}

This is a slight variation of the original AlphaZero implementation, based on recent results of \citet{efroni2018beyond}. The above equations define the planning to learning connection in Fig. \ref{fig_rtdp}. 

For the reverse connection, influencing planning based on the learned functions, we i) replace the MCTS rollout by a bootstrap estimate from the value network, and ii) modify the MCTS selects step to

\begin{equation}
\argmax_a \Bigg[ \bar{Q}(s,a) + c \cdot \pi_\theta(a|s) \cdot \sqrt{\frac{n(s,a)}{1+n(s)}} \Bigg], \label{eq_mcts}
\end{equation}

where $c \in \mathbb{R}$ is a constant that scales exploration pressure.

We vary the planning budget per timestep through adjustment of the number of traces per MCTS iteration, denoted by $n_\text{MCTS}$, while keeping the overall computational budget (in the form of wall clock time) fixed. We experiment with two well-known control tasks, CartPole and MountainCar, available from the OpenAI Gym \citep{brockman2016openai}, and with the RaceCar task, available in the PyBullet package \citep{coumans2016pybullet}. For MountainCar, we use a reward function variant with $r=-0.005$ on every step, and $r=+1$ when the Car reaches the top of the hill. Visualizations of the tasks are shown in Figure \ref{fig_tasks}. 

The total computational budget (planning, training and acting) was fixed in advance on every environment: 500 seconds for CartPole, 150 minutes for MountainCar, and 270 minutes for RaceCar. These budgets were predetermined to allow for convergence on each domain. Therefore, long planning per timestep (higher $n_\text{MCTS}$) also implies less real steps and less new training targets over the entire training period.

\begin{figure*}[h]
  \centering
      \includegraphics[width = 0.92\textwidth]{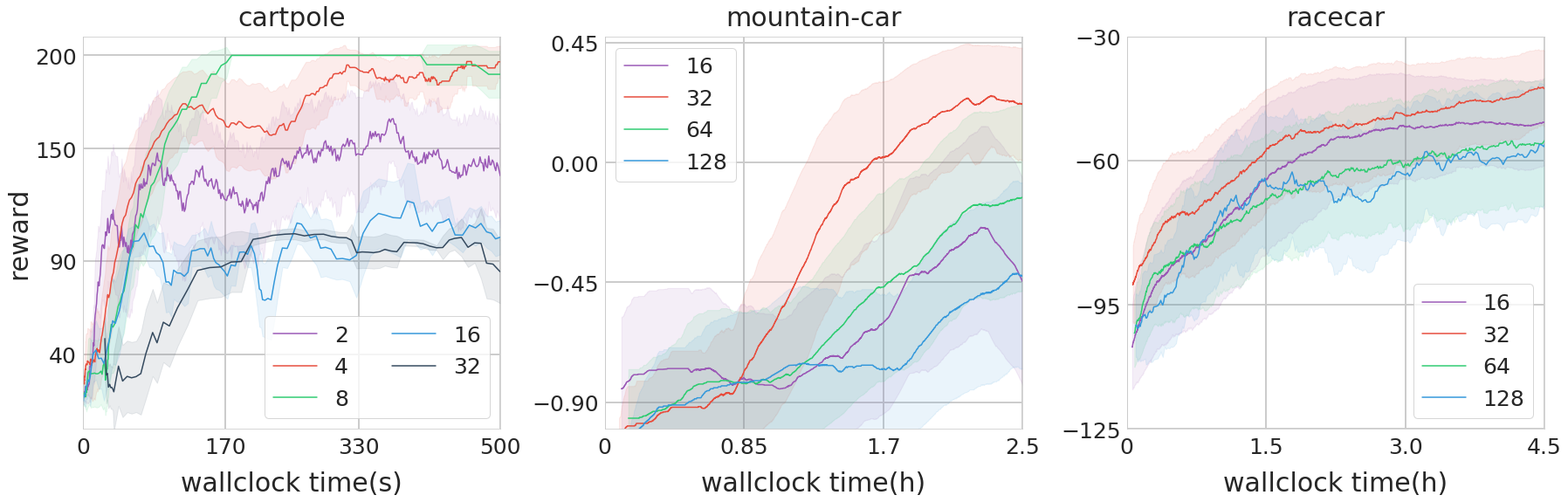}
  \caption{Learning curves on CartPole, MountainCar and RaceCar environments. The colour legend per plot displays the MCTS trace budget before every real step ($n_\text{MCTS}$). There is no clear normalization criterion for the return scales on each domain, so we report their absolute values. We see that AlphaGo Zero learns on all tasks, with best performance on CartPole, MountainCar and RaceCar achieved for budgets of, respectively, 8, 32 and 32 traces per timestep.}
    \label{fig_results_curves}
\end{figure*} 

\begin{figure*}[!t]
  \centering
      \includegraphics[width = 0.92\textwidth]{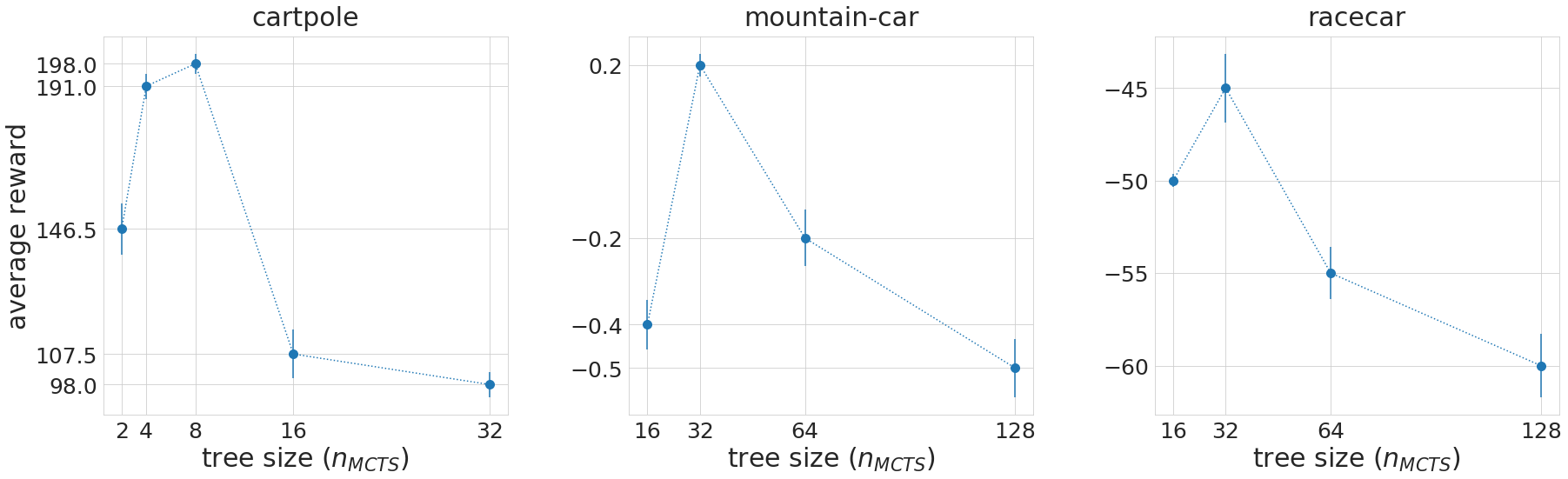}
  \caption{Trade-off between planning and learning. The horizontal axis shows the computational budget per MCTS search in the form of the total number of traces. The vertical axis shows the cumulative reward achieved by the specific set-up. Data based on last 15\% of the learning curves in Fig. \ref{fig_results_curves}. Note that the total computation time for every repetition was fixed, i.e., higher planning budget per timestep will yield less real steps and less targets for training the neural networks. We observe a clear trade-off on all domains, with optimal results achieved for intermediate search budgets.}
    \label{fig_results_tradeoff}
\end{figure*} 

\paragraph{Hyperparameters} The effect of search budget may also interact with the setting of other hyperparameters. We chose the following approach. We quickly search for a general hyperparameter configuration that shows increasing learning curves on all domains. Crucially, the search budget was varied in this quick search, but we were unaware of its actual values, to not bias the other hyperparameter settings towards good performance on a particular search budget. We will touch upon alternative approaches in the Discussion. 

We here report the fixed values for the other hyperparameters. For neural network training, we used batches of size 16 with a replay buffer of size 5e3 and learning rate of 1e-3 on all domains, optimized with ADAM optimizer \citep{kingma2014adam}. Policy and value network shared their hidden layers, with 256 hidden nodes per layer. Since the reward scales between the task varied greatly, the $c$ parameter (Eq. \ref{eq_mcts}) did require adjustment per domain: for CartPole we decayed it from 0.8 to 0.05 in 500 steps, for MountainCar from 5 to 0.5 in 5000 steps, and for RaceCar from 1.0 to 0.05 in 1500 steps. All results are averaged over 3 repetitions. 

\section{Results} \label{sec_results}
Figure \ref{fig_results_curves} shows learning curves for the three environments. We see that the AlphaZero algorithm manages to learn all three tasks. The largest variation in performance is seen on the CartPole task. Clearly, the most stable performance for CartPole uses $n_\text{MCTS}=8$. Compared to CartPole, MountainCar has a sparser reward. We therefore require longer total budget and more traces per timestep to achieve best performance, which is attained with $n_\text{MCTS}=32$. Finally, RaceCar has a larger action space than both other domains, wich requires longer training, and generally more traces per timestep. The best performance is achieved for $n_\text{MCTS}=32$ traces.

The learning curves indicate that optimal performance is achieved for an intermediate search budget. To better illustrate this observation, we aggregate the average pay-offs from the last 15\% of total time for every planning budget in each environment. These results are visualized in Figure \ref{fig_results_tradeoff}. The horizontal axis now displays search budget, while the vertical axis displays mean pay-off at the end of training. For all three environments, we observe clear optimal performance for an intermediate search budget per real step. 

\begin{figure*}[h]
  \centering
      \includegraphics[width = 0.95\textwidth]{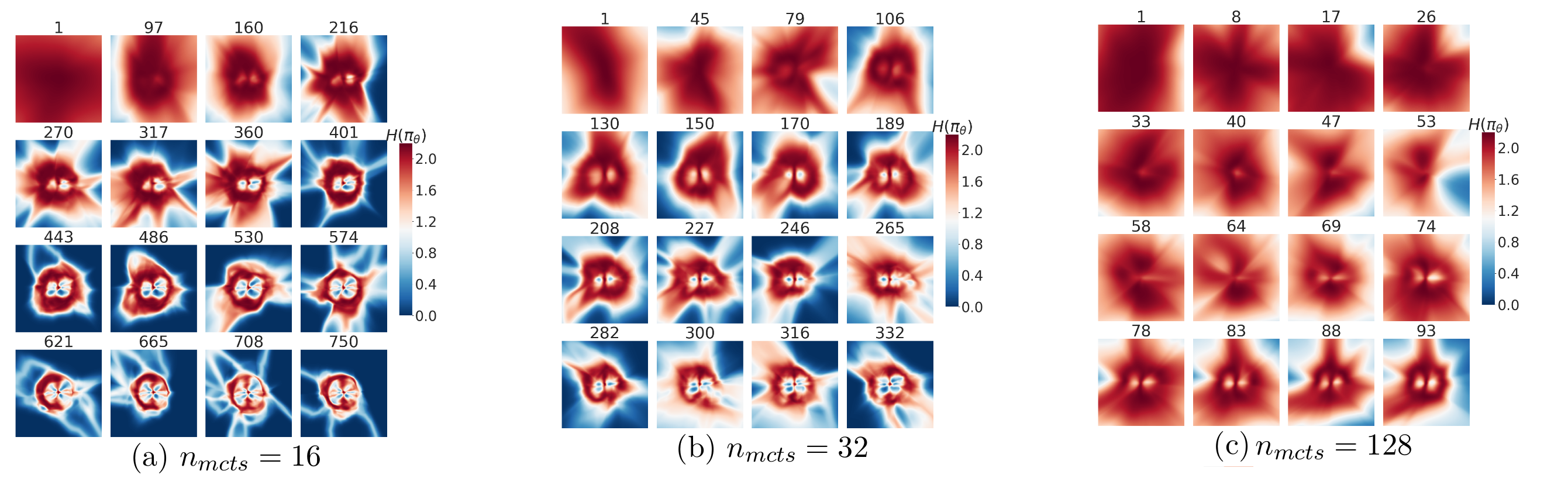}
  \caption{Training progression of policy network on RaceCar, for a) $n=16$ trace budget per MCTS iteration, b) $n=32$ trace budget, and c) $n=128$ trace budget. Each plot (a-c) visualizes a progression over training, where the number above the subplot indicates the episode number. A subplot within each plot visualizes the two-dimensional state space (x-y location of the ball in first person view), where each state is colour coded according to the entropy of the policy network at that state. High entropy (red colour) implies an uncertain policy, while low entropy (blue) implies a converged policy network. We see that the right progression ($n_\text{MCTS}=128$) qualitatively seems to slow, as there are too little training targets. The left progression ($n_\text{MCTS}=16$) seems to converge fast, but Fig. \ref{fig_results_tradeoff} shows that convergence is premature, as the achieved return is worse than the middle progression ($n_\text{MCTS}=128$).}
    \label{fig_policy_entropies}
\end{figure*} 

To further investigate what happens during training, we visualize the output of the policy network on RaceCar for different search budgets in Figure \ref{fig_policy_entropies}. The right, middle and left progression refer to $n_\text{MCTS}$ settings of 16, 32 and 128, respectively. Each subplot shows the two-dimensional RaceCar state space, which describes the (x,y)-location of the ball in first person view. Each state in this state space is coloured according to the entropy of the policy network. Red colour implies high entropy and therefore an uncertain policy, while blue colour implies low entropy and a near converged policy. The number above each subplots indicates the episode number. 

First of all, we may note that the entropy of the policy is high in the entire state space at the beginning of all three search budgets, which is to be expected. Second, we can clearly observe a difference in the number of completed episodes. Looking at the bottom-right subplot of the left ($n_\text{MCTS}=16$), middle ($n_\text{MCTS}=32$) and right ($n_\text{MCTS}=128$) plot, we observe that we completed 750, 332 and 93 full episodes for the search budgets of 16, 32 and 128 traces per real step, respectively. Of course, a higher search budget implies that we complete less episodes.

More interestingly, we can qualitatively compare the convergence of the policy networks in all three scenarios. When we compare the high search budget (right) with the intermediate one (middle), we see that the high search budget shows a similar progression, but it progresses slower. For example, the policy network at episode 93 for $n_\text{MCTS}=128$ shows similarity with the situation after episode 170 for $n_\text{MCTS}=32$, with near convergence (blue) at the border of the state space, and demarcation of early convergence areas (white) in the center of state space. Although we did require less episodes to reach that situation for $n_\text{MCTS}=128$, it did take more computation due to the relatively high planning effort per real step. Therefore, the high planning budget cannot benefit enough from generalization of information. The reverse situation is visible when we compare the left plot ($n_\text{MCTS}=16$) with the middle plot ($n_\text{MCTS}=32$). In the left plot, the policy network seems to converge faster, with a very certain policy (blue) in most of the state space at the end of the total time budget. However, if we look at the performance in Fig. \ref{fig_results_tradeoff}, the convergence was actually premature, as we probably trained on planning targets that were too unstable. We will further interpret these observations in the discussion. 

\section{Related Work} \label{sec_related_work}
AlphaGo Zero \citep{silver2017mastering} and Alpha Zero \citep{silver2018general}, as used in this work as well, are examples of multi-step approximate real-time dynamic programming. AlphaGo Zero treats the trade-off between planning and learning as a fixed hyperparameter, where they use 1600 MCTS traces per real step in the game of Go, and 800 MCTS traces per real step for both Chess and Shogi. A very similar algorithm is Expert Iteration (ExIt) \citep{anthony2017thinking}, which shows state-of-the-art performance in the game Hex. The authors do not report the MCTS budget per search used during training.

The earliest idea of iterated search and learning seems to date back to Samuel's checkers programme \citep{samuel1967some}. In later work, \citet{carmel1999exploration} explicitly studies {\it lookahead-based exploration}. The authors do mention that `it is rational for the agent to invest in computation in order to save interaction', but do not further investigate this trade-off. \citet{chang2015learning} made a step towards multi-step approximate real-time dynamic programming with Locally Optimal Learning to Search (LOLS). LOLS iterates i) Monte Carlo search, which leverages the policy, and ii) policy training, which is based on the estimated values during planning. Other algorithms that update a global value approximation based on nested search are \citet{sheppard2002world} and \citet{veness2009bootstrapping}. 

A theoretical study of multi-step greedy real-time dynamic programming was recently provided by \citet{efroni2019multi}. One of their results shows that the {\it sample} complexity of multi-step greedy RTDP scales as $\Omega(1/d)$, where $d$ denotes the depth of the lookahead, while the {\it computational} complexity scales as $\Omega(d)$. We directly see the trade-off appearing here, as deeper planning decreases the required number of real steps at the expense of increased computation. Our work provides an empirical investigation of the effect of this trade-off. Our results also seem to indicate that the optimal, intermediate planning budget also correlates with the dimensionality of the problem, where more complex problems require a higher budget. 

Our empirical results are also partly visible in the concurrent work of \citet{langlois2019benchmarking}. These authors benchmark several model-based RL algorithms. They do not focus on iterated search and RL algorithms, like multi-step approximate real-time dynamic programming, but do include results of standard RL methods that train on learned dynamics models. Their results show a similar trade-off. However, their results could also be caused by the uncertainty in a learned model, which makes planning far ahead less reliable. In contrast, our work shows a more fundamental trade-off exist, even in the case of a converged/perfect model.

As mentioned before, from a psychological perspective, our work can be related to {\it dual process theory}. Developed in the 70's and 80's by \citet{evans1984heuristic}, it describes the presence of a System 1 and System 2 in human cognition. System 1 and 2 have more recently been popularized as `Thinking Fast and Slow' \citep{kahneman2011thinking,kahneman2003maps}, respectively. System 1 includes fast, reactive, automatic behaviour, much like a neural network prediction, while System 2 includes slow, calculating, effortful decision-making, which bears similarity to local planning. This paper identifies the mutual benefit of both for optimal sequential decision making, and may as such also provide a computational motivation for the presence of both systems in humans.

\section{Discussion} \label{sec_discussion}
The computational experiments in this work clearly show a trade-off between planning, learning and acting. We identify planning budget per timestep as the major factor of importance: with a higher budget per timestep, we generate less training targets (and therefore spend less time on training) and make less real steps (complete less full episodes). 

\begin{figure}[t]
  \centering
      \includegraphics[width = 0.49\textwidth]{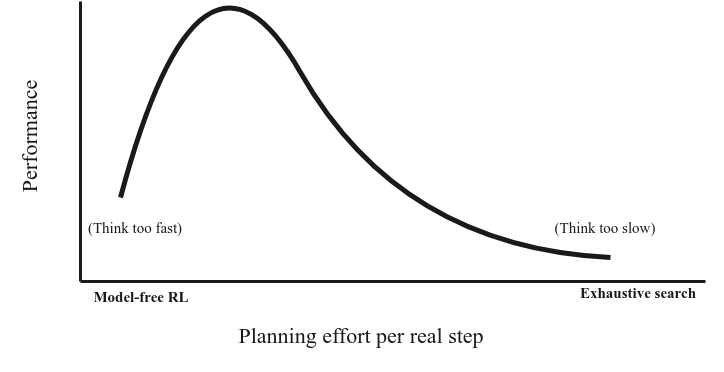}
  \caption{Conceptual illustration of the trade-off between planning and learning. The horizontal axis shows the computational budget of planning before every real step. On the left extreme we find model-free RL, which samples only a single transition before every step. On the far right, we find exhaustive search, which completely enumerates the search tree before executing a step. The curve illustrates the experimental results, which show a trade-off.}
    \label{fig_trade_off}
\end{figure}

Figure \ref{fig_trade_off} conceptually illustrates the observations from this paper. On the left of this plot, we find model-free RL, where the planning budget per timestep $\mathtt{B}=0$, and we only make real steps. Although model-free RL has shown impressive results \citep{mnih2015human}, it is known to be notoriously unstable, especially in combination with function approximation \citep{sutton2018reinforcement}. On the right of this plot we find exhaustive search, where the computational budget per timestep $\mathtt{B} \to \infty$, and we try to completely enumerate all futures from the root before choosing an action. Exhaustive search has high computational complexity that scales exponentially in the depth of the problem, and is therefore generally not a feasible approach. The problem is that it never generalizes information between states it encounters (no learning), and therefore repeats much work.

Given the above observations, the shape of Figure \ref{fig_trade_off} may come to no surprise, as it appears to keep the best of both worlds. On the one hand, we use local planning to i) create better training targets for our global value/policy approximation, and ii) correct for local errors in these approximations by looking ahead to more clearly discriminable states. On the other hand, learning adds to pure planning the ability to generalize and store global solutions in memory, which avoids repeating much work, as for example present in exhaustive search. 

As mentioned in Sec. \ref{sec_methods}, the effect of planning budget per timestep may interact with the value of other hyperparameters. For this work we chose to quickly search for a general hyperparameter setting on all domains, while being agnostic to the search budget in that phase. There could be two alternative approaches. First, we could separately optimize all other hyperparameters for every search budget on every domain. This would squeeze out the optimal performance, but is very computationally demanding. Second, we could specify an interval for every hyperparameter with reasonable values, and test on a set of random samples from these ranges, which would test robustness to hyperparameter variation. These could be interesting extensions with slightly different messages. Nevertheless, our approach is also unbiased, shows consistent results over tasks, and complies with empirical search budget decisions in other papers, for example in AlphaGo Zero \citep{silver2017mastering} (which used 1600 MCTS traces per real step, not 1 or 10 million). 

Neuroscience has suggested that both systems in dual process theory compete for control over the decision \citep{daw2005uncertainty}. Our work provides computational motivation that both systems are complementary, and actually both necessary for optimal decision making. This may also provide an evolutionary motivation for their existence.

A clear direction of future work would be to adaptively adjust the planning budget per timestep in a data-driven way. Cognitive science has for long investigated how humans decide on planning duration, aiming to find a `satisficing' (a portmanteau of satisfy and suffice) solution \citep{schwartz2002maximizing}. Computational models of such data-dependent trade-offs, possibly based on the remaining uncertainty in the plan, may further improve performance of planning-learning intergrations. 

\section{Conclusion} \label{sec_conclusion}
This paper investigated the computational trade-off between planning and learning. Our results indicate that high performance requires both local planning and global function approximation, and that the planning budget per real time-step should neither be too high nor too low. This is an important insight for the empirical application of model-based RL algorithms, but may also provide a computational motivation for the existence of a dual system in human cognition. Moreover, it opens up towards future research on this trade-off, for example identifying whether the budget per time-step should be a context-dependent function of the observed data.

\bibliographystyle{named}
\bibliography{overview}

\end{document}